\documentclass[sigconf]{acmart}
\AtBeginDocument{%
  \providecommand\BibTeX{{%
    \normalfont B\kern-0.5em{\scshape i\kern-0.25em b}\kern-0.8em\TeX}}}

\setcopyright{acmcopyright}
\copyrightyear{2018}
\acmYear{2018}
\acmDOI{10.1145/1122445.1122456}

\acmConference[Woodstock '18]{Woodstock '18: ACM Symposium on Neural
  Gaze Detection}{June 03--05, 2018}{Woodstock, NY}
\acmBooktitle{Woodstock '18: ACM Symposium on Neural Gaze Detection,
  June 03--05, 2018, Woodstock, NY}
\acmPrice{15.00}
\acmISBN{978-1-4503-XXXX-X/18/06}

\usepackage{booktabs}
\usepackage{multirow}
\usepackage{bm}
\begin{document}

\title{A Multi-view Perspective of Self-supervised Learning}

\author{Chuanxing Geng}
\affiliation{%
  \institution{Nanjing University of Aeronautics \& Astronautics}
}
\email{gengchuanxing@nuaa.edu.cn}

\author{Zhenghao Tan}
\affiliation{%
  \institution{Nanjing University of Aeronautics \& Astronautics}
}
\email{zhenghaotan@nuaa.edu.cn}

\author{Songcan Chen}
\authornotemark[1]
\affiliation{%
  \institution{Nanjing University of Aeronautics \& Astronautics}
}
\email{s.chen@nuaa.edu.cn}






%


\begin{abstract}
  As a newly emerging unsupervised learning paradigm, self-supervised learning (SSL) recently gained widespread attention in many fields such as computer vision and natural language processing. SSL aims to reduce the dependence on large amounts of annotated data in these fields, and its key is to create a pretext task without manual data annotation. Though the effectiveness of SSL has been verified in most tasks, its essence has not been well understood yet, making the design of existing pretext tasks depend more on intuitive and heuristic, thus easily resulting in some invalid presettings. In fact, from a multi-view perspective, we can find that almost all of existing SSL tasks are mainly to perform explicit linear or nonlinear transformations on the given original (single view) data, thus forming multiple artificially-generated views (i.e., view augmentation of the data). However, unlike the traditional multi-view learning, SSL mostly introduces homogeneous view data with same dimensions while exploits the views' own labels as proxy supervised signals for feature learning. As a preliminary attempt, this paper therefore borrows such a perspective to explore the essence of SSL so as to gain some insights into the design of pretext task. Specifically, we decouple a class of popular pretext tasks into a combination of view augmentation of the data (VAD) and view label classification (VLC), and design a simple yet effective verification framework to explore the impact of these two parts on SSL's performance. Our experiments  empirically uncover that it is VAD rather than generally considered VLC that dominates the performance of such SSLs.

\end{abstract}

\ccsdesc[500]{Computing methodologies~Unsupervised learning}

\keywords{Unsupervised Learning, Self-supervised Learning}


\maketitle

\section{Introduction}
Due to the powerful learning representation ability, deep convolutional neural networks (CNNs) \cite{lecun1998gradient} have made breakthroughs in such areas as computer vision and natural language processing. At present, training a powerful CNNs often requires a large number of annotated/labeled instances, however, collecting such data in numerous practical tasks is time-consuming and expensive. In contrast, unlabeled data is cheap and easily accessible. Therefore, research on the effective use of such data is already a hot issue \cite{denton2017unsupervised}. In recent years, a new learning paradigm that automatically generates supervisory signals according to some attributes of data to guide feature learning---self-supervised learning (SSL) \cite{tung2017self} is receiving increasingly more attention.

In SSL, a pretext with pseudo labels is first designed for CNNs to solve, where the pseudo labels are automatically generated based on some attributes of data without manual annotation. After the pretext task finished, the learned feature representation can be further transferred to downstream tasks as pre-trained models, making them obtain better starting points of solution \cite{jing2019self}. Thus the pretext task plays a key role, and many pretext tasks have been proposed such as in-painting \cite{pathak2016context}, patch context and jigsaw puzzles \cite{doersch2015unsupervised,noroozi2016unsupervised}, rotation \cite{gidaris2018unsupervised}, color permutation \cite{lee2019rethinking}, and so on. In particular, due to easy implementation and computational efficiency, the rotation pretext task has been more widely extended to multiple learning scenarios like domain adaptation \cite{sun2019unsupervised}, generative adversarial networks (GANs) \cite{chen2019self}, out-of-distribution detection \cite{hendrycks2019using}. Besides, as an auxiliary learning task, some researchers recently also have applied it to other learning tasks such as cross-modal learning task \cite{song2019unpaired} and face representation task \cite{rothlingshofer2019self}, etc.

\begin{figure*}[!t]
  \centering
  \includegraphics[width=0.78\textwidth]{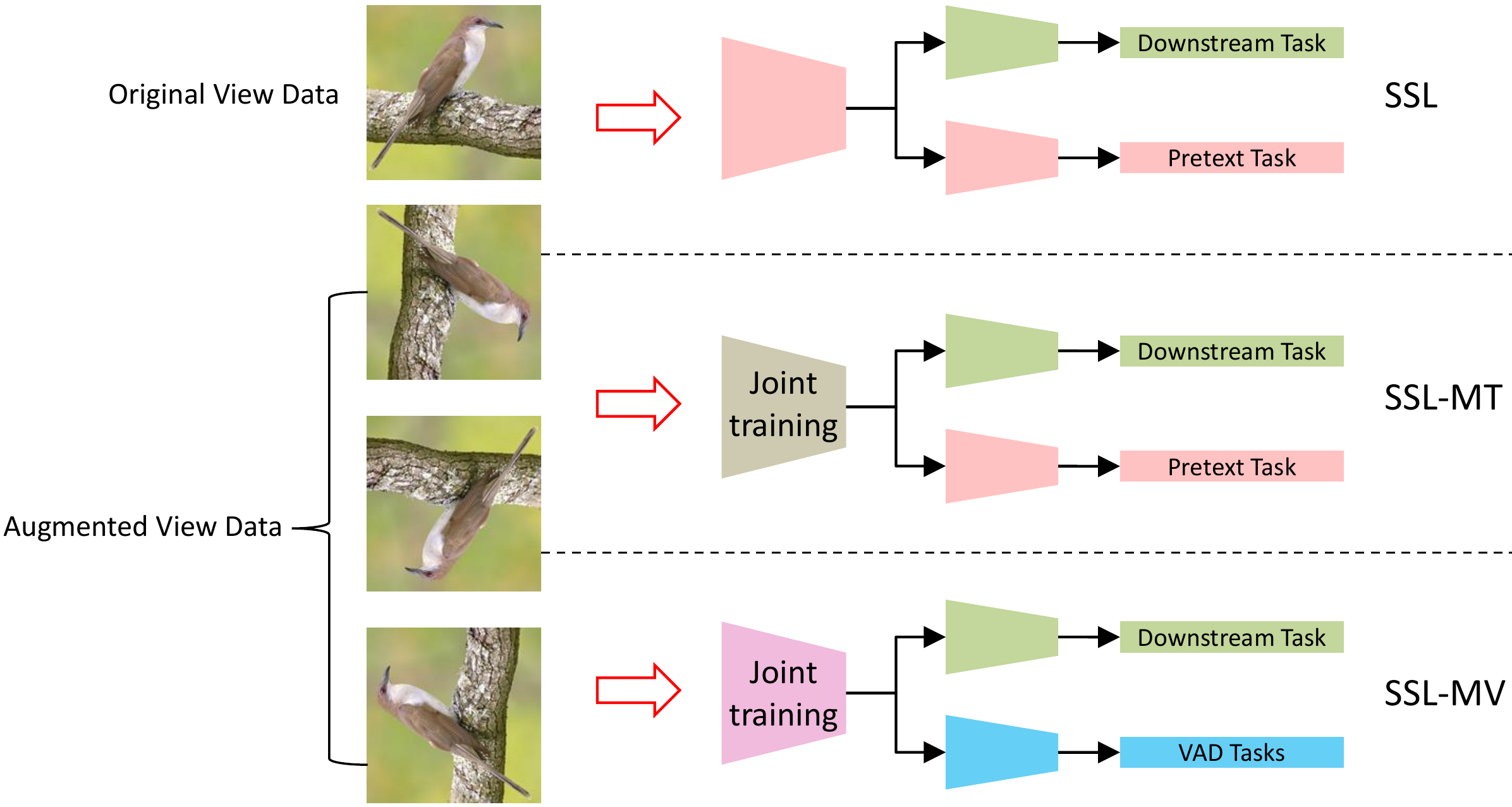}
  \caption{Comparison of different SSL implementations (taking the rotation as an example), where SSL, SSL-MT and SSL-MV denote the general, multi-task and multi-view pipelines of self-supervised learning, respectively. }
  \label{fig_sim}
\end{figure*}

Although the effectiveness of SSL has been verified in most tasks, the essence of such effectiveness has not been quite clear yet. In fact, almost all existing pretext tasks are designed based more on intuitive and heuristic, which restricts its further development. It is generally believed that the process of solving the  pretext task captures the feature representation beneficial for downstream tasks \cite{jing2019self}. However, it is a fact that the pseudo labels of pretext tasks are usually not explicitly associated with the supervised lable/signal of downstream tasks. Thus a natural question is: how does it work effectively? From a multi-view perspective, we can find that existing pretext tasks are mainly to perform explicit linear or nonlinear transformation (i.e., self-supervised signals) on the given original (single view) data, thus forming multiple transformed views (i.e., view augmentation of the data), which interestingly falls just right into the multi-view learning framework of single view data \cite{wang2007multik,wang2011novel}. \emph{However, unlike the traditional multi-view learning (MVL), SSL mainly introduces homogeneous view data with same dimension while exploits the views' own labels as proxy supervised signals to guide feature learning.} Furthermore, though some researchers \cite{federici2020learning} also conceptually involved this perspective, they have neither discriminated the difference between MVL and SSL mentioned above nor explored the essence of SSL in depth. As a preliminary attempt, this paper therefore borrows such a perspective to explore the essence so as to provide some insights into the design of pretext task.



Specifically, we first decouple a class of popular pretext tasks (e.g., rotation, color permutation) into a combination of view augmentation of the data (VAD) and view label classification (VLC). Taking the rotation as an example, such a pretext  task essentially 1) first performs a rotation transformation \{$0^\circ, 90^\circ, 180^\circ, 270^\circ$\} on the corresponding data, so as to generate multiple groups of transformed data, each group of data can be regarded as one-view new data generated from original data, all the groups form a set of so-called augmented multi-view data; 2) then predicts the rotation transformation of these data, i.e., view labels.

Further, a simple yet effective verification framework is designed to explore the impacts of VAD and VLC on SSL's performance as shown in Figure 1. Concretely, we specially design a simple multi-view learning framework for SSL (SSL-MV), which assists the feature learning of the downstream task (original view) through the same tasks on the augmented views (i.e., VAD tasks), making it focus on VAD rather than VLC. \textbf{Note that} thanks to replacing VLC with VAD tasks, SSL-MV also enables an integrated inference combining the predictions from the augmented views, further improving the performance. In addition, as a contrast, a multi-task learning framework of SSL (SSL-MT) is also provided, which focuses on both VAD and VLC. 

The rest of this paper is organized as follows. Section 2 simply reviews the recent SSL works. In Section 3, we focus on a class of popular pretext tasks (such as rotation, color permutation), and decouple them into a combination of view augmentation of the data (VAD) and view label classification (VLC) from a multi-view perspective. Then a simple yet effective verification framework is designed to help us to explore the essence of such SSL tasks. Experimental results and analyses on several benchmark datasets are reported in Section 4. Section 5 is a brief discussion, while Section 6 concludes this paper.

\begin{figure}[!t]
  \centering
  \includegraphics[width=0.5\textwidth]{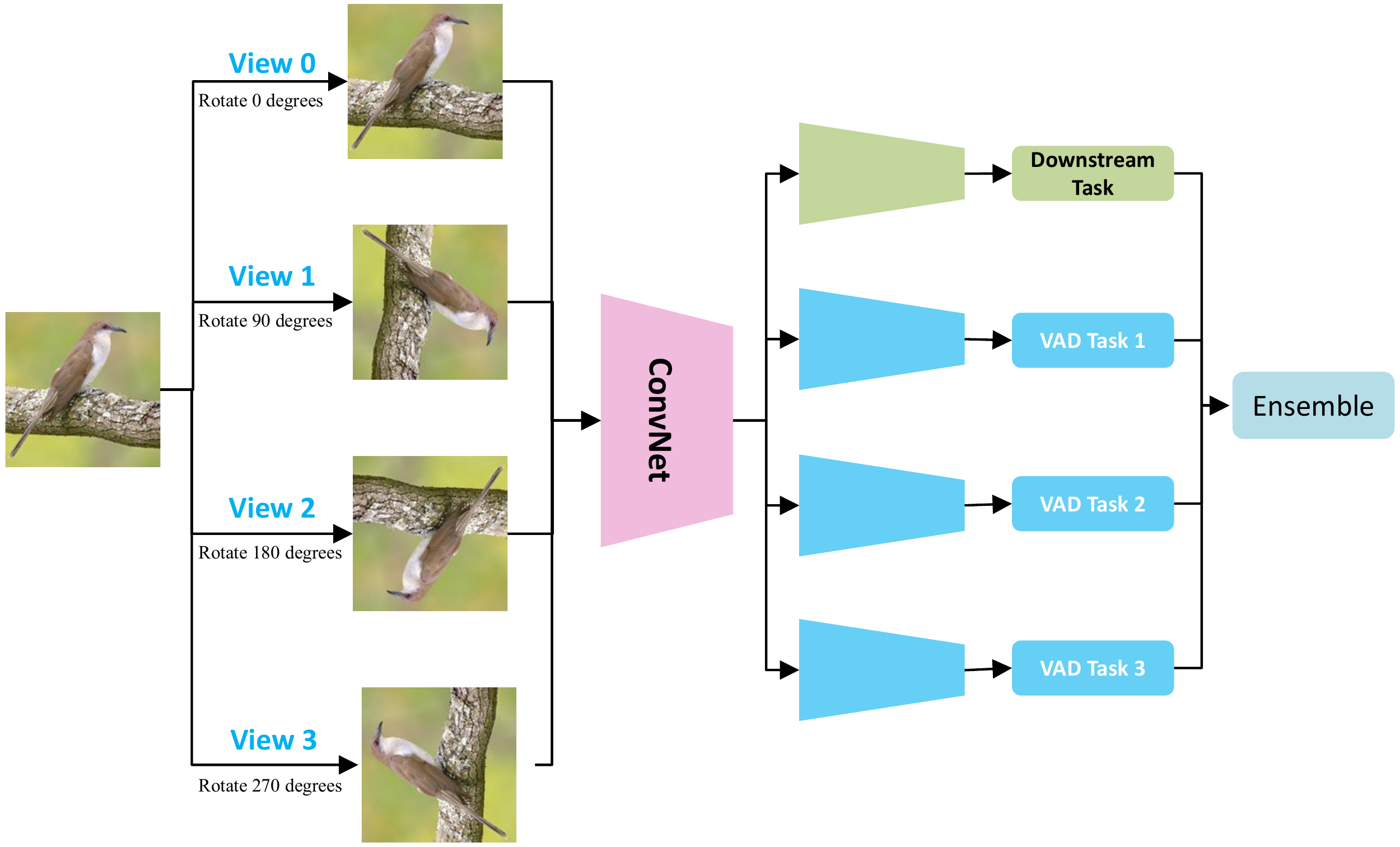}
  \caption{Overview of SSL-MV (taking the rotation as an example), where VLC in the pretext task is replaced with VAD tasks. The pink module represents the shared feature extractor $f(\bm{x})$ and the remaining network branches denote the individual classifiers $\phi_i(\cdot)$. Additionally, such a framework also enables an integrated inference combining the predictions from the different augmented views.}
  \label{fig_sim}
\end{figure}

\section{Related Work}
As the key to SSL is the design of the pretext task, much of the research currently focuses more on designing new pretext tasks, thus developing a number of SSL methods. These methods use various cues from images or videos, etc. For example, some methods learn feature representation by solving the generations involving images or videos, such as image colorization \cite{zhang2016colorful}, image inpainting \cite{pathak2016context}, and video generation with GANs \cite{tulyakov2018mocogan}, video prediction \cite{srivastava2015unsupervised}, etc. Some methods mainly adopt the image or video context information, such as jigsaw puzzles \cite{noroozi2016unsupervised}, patch context \cite{doersch2015unsupervised}, geometry \cite{dosovitskiy2015discriminative,li2019self}, counting \cite{noroozi2017representation}, clustering \cite{caron2018deep}, rotation \cite{gidaris2018unsupervised}, color permutation \cite{lee2019rethinking}, and temporal order verification \cite{misra2016shuffle,wei2018learning}, sequence prediction \cite{lee2017unsupervised}, etc. Furthermore, there are also some methods related to the automatic generation of free semantic label by traditional hardcode algorithms \cite{faktor2014video} or by game engines \cite{ren2018cross}, such as semantic segmentation \cite{wang2020self}, contour detection \cite{ren2018cross}, relative depth prediction \cite{jiang2018self}, etc. For more details about these methods mentioned above, we refer the reader to \cite{jing2019self}.

\begin{table}
\centering
\tabcolsep 1.5mm
\renewcommand\arraystretch{1.2}
\caption{Statistics of the four benchmark datasets.}
\begin{tabular}{lccc}
\toprule
\textbf{Dataset}&      \textbf{\# Class} &\textbf{\# Training} &\textbf{\# Testing} \\
\midrule
 CIFAR10      & 10     &50000       & 10000      \\
 CIFAR100     & 100    &50000       & 10000       \\
 TinyImageNet & 200    &100000      & 10000       \\
 MIT67        & 67     &5360        & 1340        \\
\bottomrule
\end{tabular}
\label{tab:booktabs}
\end{table}

In addition to exploring the design of pretext tasks, some researchers have also attempted to analyze the inherent advantages of SSL from other perspectives. From the perspective of CNNs' structure, Kolesnikov et al. \cite{kolesnikov2019revisiting} studied the combination of multiple CNNs' structures and multiple SSL pretext tasks, where they found that unlike supervised learning, the performance of SSL tasks depends significantly on the structure of CNNs used. Furthermore, they also experimentally show that the performance advantages obtained by the predecessor tasks do not always translate into the advantages in the feature representation of the downstream tasks, sometimes even worse. From the robust learning perspective, Hendrycks et al. \cite{hendrycks2019using} found that SSL can improve robustness in a variety of ways, such as robustness to adversarial examples, label corruption, common input corruptions, while it also greatly benefits out-of-distribution detection \cite{hendrycks2016baseline,lee2017training} on difficult, near-distribution outliers.

From the data augmentation perspective, Lee et al. \cite{lee2019rethinking} reorganized various SSL strategies, where they mainly regarded SSL as a data augmentation technique and modeled the joint distribution of the original and self-supervised labels. Further, Federici et al. \cite{federici2020learning} empirically demonstrated the effectiveness of this kind of data augmentation under the multi-view information bottleneck framework they proposed. Besides, Keshav and Delattre \cite{keshav2020self} explored the crucial factors to avoid learning trivial solutions when constructing pretext tasks, while Kumar et al. \cite{kumar2020data} investigated the SSL task combined with clustering, and provided the insight from data transformation perspective.


\section{Proposed Frameworks}
In this part, we focus on a class of popular pretext tasks such as rotation and color permutation, where we decouple them into a combination of view augmentation of the data (VAD) and view label classification (VLC), so as to explore whether VAD or VLC dominates the performance of such SSL. Specifically, two implementation pipelines of SSL, i.e., SSL-MT and SSL-MV, are designed, respectively, where SSL-MT focused on both VAD and VLC, while SSL-MV concentrates only on VAD. The more details about them will be provided in the following parts.

%

\textbf{Notation.} Let $\mathcal{D}=\{(\bm{x}_i,y_i)\}_i^N\subseteq \mathbb{R}^{N\times d}$ denote the training set, where $y_i\in\{1,2,...,K\}$ is the corresponding label and $K$ is the number of classes. Let $f(\bm{x})$ represent the shared feature extractor, $\mathcal{L}_{\text{CE}}$ be the cross-entropy loss function, and $\phi_i(\cdot)$ be the individual classifier. Furthermore, let $\bm{x}^{(j)}$ denote the augmented sample using a transformation $T_j$ (self-supervision label), where $j = 0,1,2,...,M$, and $T_0$ denotes the \emph{identical transformation}.

\begin{figure*}[!t]
  \centering
  \includegraphics[width=0.78\textwidth]{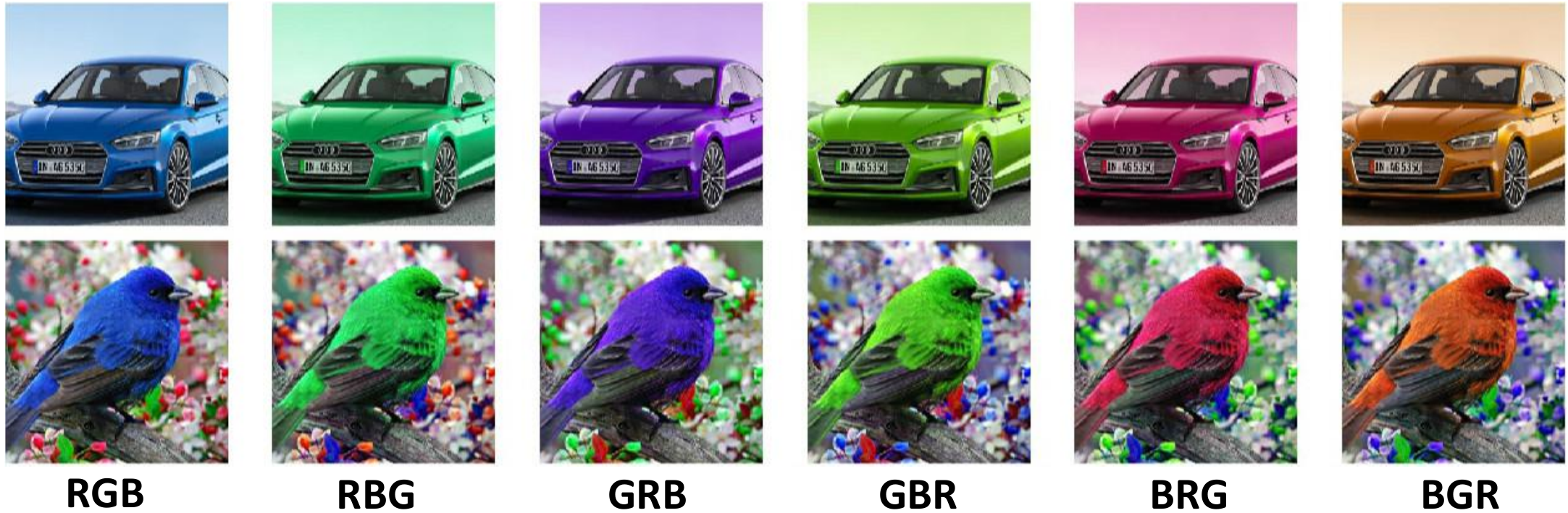}
  \caption{Visualization of color permutation instances.}
  \label{fig_sim}
\end{figure*}

\subsection{Multi-task pipeline of SSL}
Most existing SSL methods usually adopt the two-step strategy, where they first solve a pretext task, i.e., predicting which transformation $T_j$ (self-supervised/view label) is applied to the original sample $\bm{x}$, then transfer the learned feature representation to downstream tasks. However, since the self-supervised label is usually not explicitly associated with the supervised label of downstream tasks, thus the performance of such implementation pipeline currently lags behind fully supervised training, as reported in recent literature \cite{gidaris2018unsupervised,hendrycks2019using}. Therefore, to better explore the essence of SSL, we here adopt a multi-task pipeline to implement SSL. Specifically, we formulate the loss function as follows,
\begin{eqnarray}
L_{\text{MT}}(f,\bm{\phi}) &=& \frac{1}{N}\sum_{i=1}^N\mathcal{L}_{\text{CE}}(\phi_{\text{down}}(f(\bm{x}_i^{(0)})),y_i) + \\ \nonumber &&\sum_{j=0}^M\mathcal{L}_{\text{CE}}(\phi_{\text{pret}}(f(\bm{x}_i^{(j)})),j),
\end{eqnarray}
where the first term of the RHS of the equality (1) denotes the downstream task loss, while the second term is the pretext task loss. Many recent SSL works almost all adopted such a pipeline \cite{sun2019unsupervised,chen2019self}.

\begin{figure}[!t]
  \centering
  \includegraphics[width=0.48\textwidth]{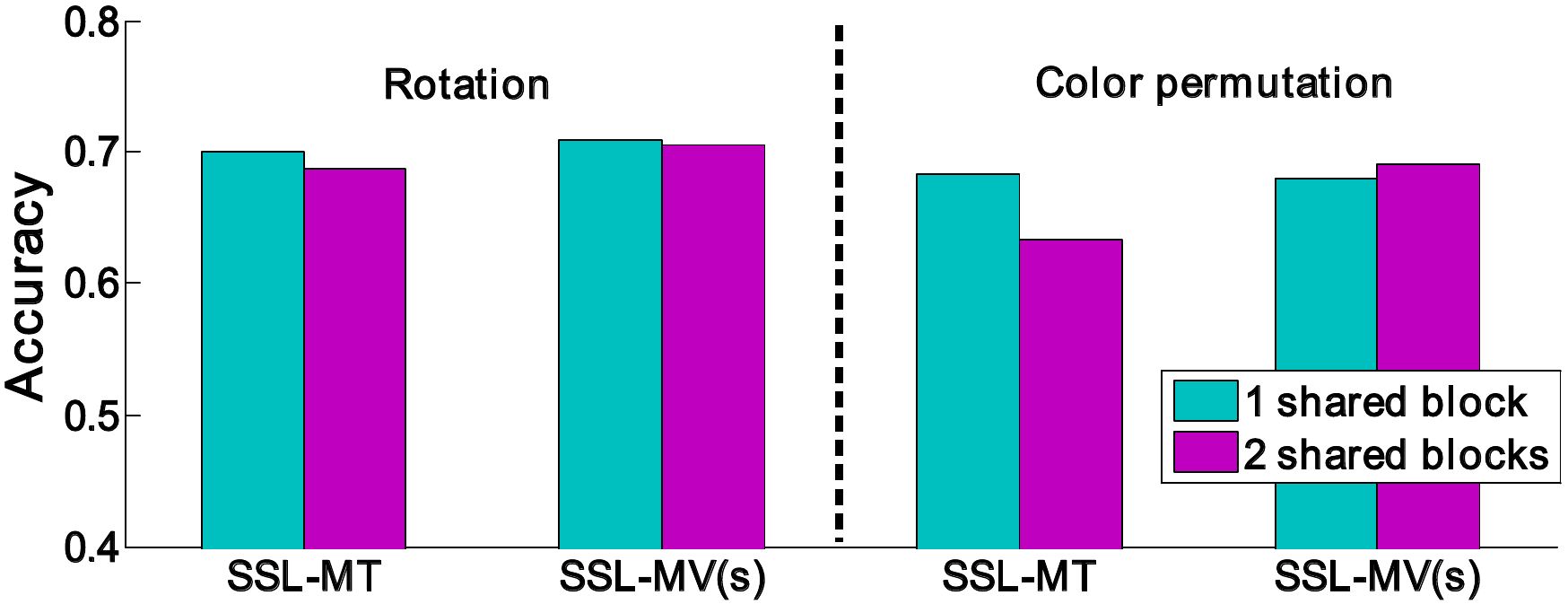}
  \caption{Shared block experiment on CIFAR100. }
  \label{fig_sim}
\end{figure}

\subsection{Multi-view pipeline of SSL}
To effectively investigate whether VAD or VLC dominates SSL's performance, we specially design a simple yet effective multi-view learning framework for SSL as shown in Figure 2, which assists the feature learning of downstream tasks (original view) through the same supervised learning task on the augmented views (VAD Tasks). Further, we define the following loss function:
\begin{equation}
L_{\text{MV}}(f,\bm{\phi}) = \frac{1}{N}\sum_{j=0}^{M}\sum_{i=1}^{N}\mathcal{L}_{\text{CE}}(\phi_j(f(\bm{x}_i^{(j)})),y_i).
\end{equation}

Different from SSL-MT which focuses on both VAD and VLC, SSL-MV only pays attention to VAD. Furthermore, from the multi-task perspective, SSL-MV essentially replaces the VLC in SSL-MT with VAD tasks. \textbf{Note that}, intuitively, such tasks are more helpful for the feature learning of downstream tasks, since they are closer to the downstream task compared to the VLC tasks. Additionally, SSL-MV actually also enables an integrated inference combining the predictions from the different augmented views, consequently further improving the performance  as validated in our experiments below.

\section{Experiments and Analysis}
\subsection{Settings}
\textbf{Datasets and models.} In this paper, we evaluate our learning framework on several commonly used benchmark datasets: CIFAR10/100 \cite{krizhevsky2009learning}, Tiny-ImageNet (https://tiny-imagenet.herokuapp.com/), and a fine-grained dataset MIT67 \cite{quattoni2009recognizing}. Moreover, Table 1 describes the details of these datasets. For the CNN models used, we refer to \cite{lee2019rethinking}: 32-layer Reset for CIFAR, 18-layer ResNet for Tiny-ImageNet and MIT67.

\textbf{Training Details.}
For all benchmark datasets, we use SGD with learning rate 0.1, momentum 0.9. For CIFAR10/100, we set weight decay to $5\times 10^{-4}$, and train for 200 epoches with batch size of 128. For Tiny-ImageNet, we set weight decay of $10^{-4}$, and train for 100 epoches with batch size of 64 (limited by the GPU capacity). For MIT67, we set weight decay of $5\times 10^{-4}$, and train for 180 epoches with batch size of 32 as suggested by \cite{lee2019rethinking}. We decay the learning rate by the constant factor of 0.1 at 50\% and 75\% epoches. Furthermore, for SDA+AI, SDA+AG and the standard supervised learning, we follow the same settings in \cite{lee2019rethinking}.

\textbf{Choices of pretext tasks.} As we use image classification as the downstream task, where using the entire image as input is important. Therefore, similar to \cite{lee2019rethinking}, we here focus on two pretext tasks of such SSL, i.e., rotation and color permutation. For rotation, we adopt four commonly used transformations, i.e., $\{0^{\circ},90^{\circ},180^{\circ},270^{\circ}\}$, while six transformations for color permutation, i.e., \{RGB, RBG, GRB, GBR, BRG, BGR\} as visualized in Figure 3. \emph{Note that though we here mainly carry out experimental studies based on both rotation and color permutation, some drawn conclusions can give insights of the essence of SSL.}

\subsection{Shared block experiment}
Generally, the shallow layers of CNNs can capture general low-level features like edges, corners, etc, while deeper layers capture task-related high-level features \cite{jing2019self}. This subsection reports the impact of sharing different numbers of blocks on performance. Taking CIFAR100 as an example, Figure 4 shows the corresponding results. For SSL-MT, its performance is negatively correlated with the number of shared blocks, which is consistent with the previous general conclusion. However, for SSL-MV(s), its performance of sharing two blocks is at least equal to that of sharing one block, which is probably because the VAD tasks in SSL-MV are closer to downstream tasks than the VLC task in SSL-MT. Additionally, for color permutation, though sharing more layers improves the performance of SSL-MV(s) (67.99\% vs 69.09\%), on the other hand, the performance of SSL-MV(e) decreases (76.32\% vs 75.11\%). For the sake of fairness, all experiments in this paper share the first block of the corresponding CNNs.
\begin{table*}[]
\centering
\tabcolsep 2.3mm
\renewcommand\arraystretch{1.2}
\caption{Classification accuracy (\%) using rotation and color permutation. SSL-MV(s) and SSL-MV(e) respectively denote the single inference and aggregated inference of SSL-MV. '-' indicates the corresponding method does not provide their results, while the best accuracy is indicated as bold.}
\begin{tabular}{c|c|cccc|cc}
\hline
\multicolumn{2}{c|}{\textbf{Dataset} \textbackslash \textbf{Method}} & Supervised & SDA+SI   & SSL-MT & SSL-MV(s) & SDA+AG & SSL-MV(e) \\ \hline
CIFAR10      & \multirow{4}{*}{\textbf{Rotation}} & 92.39     & 92.50   & 93.81 & 93.47 & 94.50    & \textbf{95.18}    \\
CIFAR100     &                                    & 68.27     & 68.68   & 70.11 & 70.89 & 74.14    & \textbf{77.26}    \\
TinyImageNet &                                    & 63.11     & 63.99   & 64.08 & 64.37 & 66.95    & \textbf{69.51}    \\
MIT67        &                                    & 54.75     &  -       & 59.18 & 58.28 & \textbf{64.85}    & 61.49    \\ \hline
CIFAR10      & \multirow{4}{*}{\textbf{Color permutation}}    & 92.39     &  -       & 92.42 & 93.52 & 92.51    & \textbf{95.01}    \\
CIFAR100     &                                                & 68.27     &  -       & 68.30 & 67.99 & 69.14    & \textbf{76.32}    \\
TinyImageNet &                                                & 63.11     &  -       & 64.47 & 64.25 & 64.15    & \textbf{70.51}    \\
MIT67        &                                                & 54.75     &  -       & 54.93 & 57.24 & 59.99    & \textbf{60.30}     \\ \hline
\end{tabular}
\end{table*}

\subsection{Main Results}
To investigate whether whether VAD or VLC dominates the performance of SSL, we here focus most on the comparison between SSL-MT and SSL-MV. In addition, for SSL-MV, it has a single inference (SSL-MV(s)) and an aggregated inference (SSL-MV(e)). Thus in order to further demonstrate the advantages of SSL-MV, we also respectively report their comparison with SDA + SI (a single inference) and SDA + AG (an aggregated inference) in \cite{lee2019rethinking}, a latest work closed to our SSL-MV. Moreover, we report the results of the standard supervised learning as a baseline. Table 2 shows the results.

\textbf{Observation 1:} Compared to the standard supervised learning, SSL-MT and SSL-MV achieve significant performance improvements on almost all datasets, which demonstrates that such pretext tasks indeed benefit the feature learning for the downstream tasks.

\textbf{Observation 2:} No matter whether it is rotation or color permutation, SSL-MV(s) achieves at least comparable performance to SSL-MT on all benchmark datasets, which fully indicates that it is VAD rather than generally considered VLC that dominates the performance of such SSL, since it has been replaced by the VAD tasks. This also inspires us to further improve the performance of SSL by generating richer data views, such as using transform composition, which will be discussed in subsection 4.5

\begin{table*}[]
\centering
\tabcolsep 3.7mm
\renewcommand\arraystretch{1.05}
\caption{Classification accuracy (\%) of color permutation experiments on CIFAR100.}
\begin{tabular}{ccccc}
\hline
\multirow{2}{*}{\textbf{Color permutation}} & \multirow{2}{*}{\textbf{Views}}&\multirow{2}{*}{\textbf{SSL-MT}} & \multicolumn{2}{c}{\textbf{SSL-MV}} \\ \cline{4-5}
                       &                        &      & SSL-MV(s)     & SSL-MV(e)    \\ \hline
RGB                    & 1                      &68.27& \multicolumn{2}{c}{68.27}   \\
RGB,GBR,BRG            & 3                      &69.54& 66.99        & \textbf{73.84}       \\
RGB,GRB,BGR            & 3                      &69.24& 69.69        & \textbf{73.89}       \\
RGB,RBG,GRB,GBR,BRG,BGR        & 6              &68.30& 67.99        & \textbf{76.32}       \\ \hline
\end{tabular}

\end{table*}

\textbf{Observation 3:} Compared to the single inference, our aggregated inference (SSL-MV(e)) achieves the best performance on all datasets. This demonstrates that the transforms from such pretext tasks indeed enrich the diversity of the original data, which in turn leads to the diversity of the classifiers, consequently further improving the performance \cite{krawczyk2017ensemble}. Besides, both SSL-MV(s) and SSL-MV(e) are significantly ahead of SDA+SI and SDA+AG, respectively, on almost all datasets. This seems to validate our previous conjecture, i.e., there may be potentially unfavorable competition between the original and self-supervised labels in such a modeling of \cite{lee2019rethinking}. Note that, although the VAD tasks also compete with the downstream tasks,  the resulting competition has eased due to the closeness between these tasks.

\textbf{Observation 4:} The performance of SSL-MV(s) lags behind the standard supervised learning on CIFAR100, and we conjecture that some view(s) of VAD may confuse the shared feature extractor. In other words, some augmented view(s) may introduce severe distribution drift. As shown in Figure 3, taking the bird as an example, the newly generated GBR and BRG images have visually shifted greatly from the original RGB images. Especially, when these images have largely color overlap with other kinds of birds, it is easy to confuse the classifiers. To verify our conjecture, we further conduct the color permutation experiments in subsection 4.4.

\textbf{Remark.} Note that, this paper focuses on the exploration of the essence of SSL, and to better study this issue, we specially design such a simple yet effective multi-view learning framework for SSL. Although the single inference, especially aggregated inference of this framework has achieved impressive performance as reported above, this framework currently is not scalable and difficult to be directly deployed into practical applications.

\begin{table*}[]
\centering
\tabcolsep 2.8mm
\renewcommand\arraystretch{1.05}
\caption{Classification accuracy (\%) depending on the set of transformations.}
\begin{tabular}{ccccc}
\hline
\multirow{2}{*}{\textbf{Rotation}} & \multirow{2}{*}{\textbf{Sharpness}} & \multirow{2}{*}{\textbf{Views}} & \multicolumn{2}{c}{\textbf{SSL-MV}} \\ \cline{4-5}
                          &                            &                        & SSL-MV(s)     & SSL-MV(e)    \\ \hline
        $0^\circ$         & $\gamma=1$                        & 1                      & \multicolumn{2}{c}{68.27}   \\
  $0^\circ, 180^\circ$    & $\gamma=1$                        & 2                      & 70.87        & \textbf{75.08}       \\
  $0^\circ, 90^\circ, 180^\circ, 270^\circ$& $\gamma=1$       & 4                      & 70.89        & \textbf{77.26}       \\
          $0^\circ$       & $\gamma=0,0.5,1,1.5$              & 4                      & 69.36        & \textbf{75.87}      \\
  $0^\circ, 180^\circ$ & $\gamma=0,0.5,1,1.5$    & 8              & 69.39        & \textbf{77.69}       \\
  $0^\circ, 90^\circ, 180^\circ, 270^\circ$ & $\gamma=0,0.5,1,1.5$    & 16             & 69.17        & \textbf{77.72}       \\ \hline
\end{tabular}

\end{table*}

\subsection{Color permutation experiment}
In this subsection, we specifically perform the color permutation experiments to verify our claims in subsection 4.3, and the results are reported in Table 3. As we conjectured, SSL-MV(s) lags behind the standard supervised learning when the augmented views GBR and BRG are used to assist the feature learning of the downstream task. In contrast, when GRB and BGR are used, the performance of SSL-MV(s) is significantly better than that of the standard supervised learning, indicating that the distribution drift on the latter is smaller than the former. Similarly, SSL-MV(s) performs slightly better than SSL-MT when using GRB and BGR, whereas it lags behind SSL-MT when using GBR and BRG, which also indicates that VLC is necessary when the samples are sensitive to the corresponding transformation (color permutation here). Interestingly, though the distribution drift of some augmented view data interferes with the feature learning of downstream tasks to some extent, the performance of aggregated inference (SSL-MV(e)) is positively correlated with the number of augmented views.

\subsection{Composed transformations}
As mentioned in subsection 4.3, VAD dominates the performance of such SSL discussed in this paper, which inspires us to further improve
the performance of SSL by generating richer data views, such as through transform composition. As a preliminary exploration, an extended experiment on the composed transformations containing rotation and an image enhancement transformation, i.e., $\text{Sharpness}(\gamma)$, is carried out in this subsection. The parameter $\gamma$ controls the sharpness of an image, where $\gamma=0$ returns a blurred image, while $\gamma=1$ returns the original image. Specifically, we adopt $\gamma=0,0.5,1,1.5$ four transformations, and the corresponding results are reported in Table 4.

For SSL-MV(s), although its performance improves with the increase of augmented view data, it is not positively correlated to the number of the augmented views. One possible reason is that some augmented views of VAD drift too much and confuse the shared feature extractor as discussed above. Nonetheless, such augmented view data greatly increase the diversity of the original data, making our aggregated inference (SSL-MV(e)) achieve better performance.

\section{Discussion}
From the multi-view perspective, this paper revisits a class of popular SSL methods as an attempt to preliminarily explore the essence of SSL, where we find it is VAD rather than generally considered VLC plays a crucial role in boosting performance for downstream tasks. Please note that unlike traditional data augmentation and multi-view learning, SSL also considers the so-attached self-supervised labels. Furthermore, to our best knowledge, the theoretical exploration of SSL currently is almost blank as well. Thus, from the above explorations, it seems that we can draw on the existing multi-view learning theory \cite{wang2017theoretical} to compensate this issue, and further expand SSL by the following typical strategies:
\begin{enumerate}
\item Generate more diverse self-supervised signals/labels. For example, through richer transformation compositions: composed/nested/hierarchical operations on data examples.
\item Establish a unified learning framework of  self-supervised learning and multi-view learning.
\end{enumerate}
Additionally, inspired by the effective use of self-supervised signals in SSL, exploring the use of view labels in existing multi-view learning is also worth further in-depth discussion.

\section{Conclusion}
In this paper, we revisit SSL from the multi-view perspective as an attempt to explore its essence, where we decouple a class of popular pretext tasks into a combination of view data augmentation (VAD) and view label classification (VLC), empirically revealing that it is VAD rather than generally considered VLC that dominates the performance of such SSL. Furthermore, we also give some insights about SSL, hoping to benefit the researchers in this community.

\clearpage
\bibliographystyle{ACM-Reference-Format}
\bibliography{acm2020}

%
%
%
%
%
%
%
%

\end{document}